\documentclass{article} 

\usepackage{colm2024_conference}

\usepackage{microtype}
\usepackage{hyperref}
\usepackage{url}
\usepackage{booktabs}
\usepackage{amsmath}
\usepackage{amsfonts}
\usepackage{amssymb}
\usepackage[pdftex]{graphicx}

\usepackage{adjustbox}
\definecolor{darkblue}{rgb}{0, 0, 0.5}
\hypersetup{colorlinks=true, citecolor=darkblue, linkcolor=darkblue, urlcolor=darkblue}
\usepackage{array}
\usepackage[utf8]{inputenc}
\usepackage{multirow} 
\usepackage{hhline} 
\usepackage{array} 
\usepackage{colortbl} 
\usepackage{caption} 
\usepackage{tabularx}
\usepackage{enumitem}

\title{QANA: LLM-based Question Generation and Network Analysis for Zero-shot Key Point Analysis and Beyond}

\author{Tomoki Fukuma, Koki Noda, Toshihide Ubukata\& Kousuke Hosoi
 \\
TDAI Lab Co., Ltd.\\
Tokyo, Japan\\
\texttt{\{tomoki.fukuma, koki.noda, toshihide.ubukata, kousuke.hosoi\}@tdailab.com} \\
\And
Yoshiharu Ichikawa, Kyosuke Kambe, \& Yu Masubuchi \\
NHK (Japan Broadcasting Corporation) \\
Tokyo, Japan \\
\texttt{\{ichikawa.y-gq,kambe.k-je, masubuchi.y-lq\}@nhk.or.jp} \\
\AND
Fujio Toriumi \\
the University of Tokyo \\
Tokyo, Japan \\
\texttt{tori@sys.t.u-tokyo.ac.jp}
}

\colmfinalcopy 
\begin{document}

\maketitle

\begin{abstract}
The proliferation of social media has led to information overload and increased interest in opinion mining. We propose "Question-Answering Network Analysis" (QANA), a novel opinion mining framework that utilizes Large Language Models (LLMs) to generate questions from users' comments, constructs a bipartite graph based on the comments' answerability to the questions, and applies centrality measures to examine the importance of opinions. We investigate the impact of question generation styles, LLM selections, and the choice of embedding model on the quality of the constructed QA networks by comparing them with annotated Key Point Analysis datasets. QANA achieves comparable performance to previous state-of-the-art supervised models in a zero-shot manner for Key Point Matching task, also reducing the computational cost from quadratic to linear. For Key Point Generation, questions with high PageRank or degree centrality align well with manually annotated key points. Notably, QANA enables analysts to assess the importance of key points from various aspects according to their selection of centrality measure. QANA's primary contribution lies in its flexibility to extract key points from a wide range of perspectives, which enhances the quality and impartiality of opinion mining.
\end{abstract}

\section{Introduction}\label{introduction}
The proliferation of social media has led to information overload and increased interest in opinion mining as a means to understand the main points of online discussions. Automatic summarization, a major approach in this field (\cite{angelidis-lapata-2018-summarizing}), has limitations such as the lack of quantitative aspects, inability to indicate the prevalence of each point (\cite{bar-haim-etal-2020-arguments}), and potential issues like hallucination (\cite{maynez-etal-2020-faithfulness}), fairness (\cite{zhang2023fair}), or the extraction of a limited set of opinions that may not fully represent the diverse viewpoints on controversial topics (\cite{Huang-et-al-NAACL-24}).

Key Point Analysis (KPA) has been proposed as an alternative to address these limitations (\cite{bar-haim-etal-2020-arguments,bar-haim-etal-2020-quantitative}). KPA consists of two subtasks: Key Point Generation (KPG), which generates multiple short sentences describing main points from a corpus, and Key Point Matching (KPM), which maps each argument to the appropriate key points. Unlike traditional summarization, KPA quantifies the prevalence of each key point by the number of matching sentences, providing insight into the prominence of each point.

However, in the broadcasting domain, topic selection plays a crucial role, and using auto-generated summaries or high-volume key points undoubtedly is unacceptable according to agenda-setting theory (\cite{McCombsShaw1972}). Organizations like the BBC prioritize impartiality and fairness (\cite{flood2011between}), emphasizing the need for diverse viewpoints. Therefore, an analytical framework that evaluates the significance of key points from multiple perspectives beyond mere participation is essential.

To address this issue, we propose a novel opinion mining framework called "Question-Answering Network Analysis" (QANA). Our approach involves generating questions from original comments using Large Language Models (LLMs) and constructing a Question-Answering (QA) network based on each argument's ability to answer those questions. Consequently, various centrality measures (\cite{Lu2016VitalNodes}) can be applied to identify important nodes, enabling the evaluation of key point significance from diverse perspectives beyond mere quantitative aspects. For instance, centrality measures such as PageRank (\cite{page1999pagerank}) or betweenness (\cite{Freeman1977Set}) can capture the structural importance of key points within the overall discussion, considering factors beyond simple metrics like degree centrality.

In this study, we evaluate the quality of networks generated through our methodology by comparing them with human-annotated datasets in the KPA experimental setting. The evaluation involves altering the question style and the choice of LLMs. First, through KPM, we assess whether arguments and annotated key points correspond via questions within the network. Second, we examine if questions with high centrality align with manually crafted topics in the KPG setting.

The main contributions of this study are: (1) In the KPM task, the QANA approach achieved performance comparable to traditional supervised learning models, even in a zero-shot setting. (2) The computational cost of matching key points and arguments was reduced from quadratic to linear. (3) In the KPG task, our approach demonstrated that questions with high PageRank or degree centrality align well with manually crafted key points. (4) The primary contribution is the proposal of a flexible framework that allows analysts to assess the importance of key points from various aspects according to their interests, providing a valuable tool for opinion mining in the broadcasting field where impartiality and fairness are crucial considerations.

\section{Related Work}

\subsection{Key Point Analysis}\label{kpa}

Key Point Analysis (KPA) is a task proposed by \cite{bar-haim-etal-2020-arguments} that consists of two main steps: "Key Point Generation" (KPG) and "Key Point Matching" (KPM).

Key Point Generation (KPG) is a type of Multi-Document Summarization (MDS) that aims to generate discussion topics. Unlike conventional summarization methods, KPG generates multiple short sentences, representing the key points of the discussion. Previous attempts have utilized neural topic modeling to cluster similar opinions and employed generative summarization techniques (\cite{li-etal-2023-hear,van-der-meer-etal-2024-empirical}). The difficulty in KPG lies in evaluating the key point sets, which differs from traditional summarization evaluation. In KPG, the evaluation must consider not only the similarity between individual sentences but also the quality of the key points as a set, capturing the diverse viewpoints present in the discussion. This aspect has not been thoroughly investigated until recently. \cite{li-etal-2023-hear} proposed a new evaluation framework that measures the similarity between generated topics and manually created key points using metrics such as BERTScore to calculate soft precision and soft recall. However, this approach requires ground truth key points and cannot provide a multifaceted evaluation of the generated results, which has been discussed in recent reference-free summarization metrics(\cite{liu-etal-2022-reference}). This study also introduces a novel approach from an evaluation perspective, utilizing centrality measures for the set of key points in KPG. This allows for a diverse range of evaluations on the key points, making it a new attempt in the field of KPG evaluation.

Key Point Matching (KPM) aims to match arguments to their corresponding key points based on stance agreement. Most KPM methods employ Transformer-based models like RoBERTa (\cite{liu2019roberta}) to predict the agreement between each comment-key point pair by inputting them together (\cite{friedman-etal-2021-overview}). However, the computational cost of these pair-wise comparisons becomes a bottleneck for large datasets (\cite{eden-etal-2023-welcome}), leading to challenges in the subsequent computational complexity. To address this issue, various approaches have been proposed, such as limiting the number of key point candidates and selecting key points based on a subset of sentences (\cite{eden-etal-2023-welcome}). Despite these improvements, matching run time remains a major challenge when deploying a KPA system that should serve many users and process large datasets.

\subsection{Exploring Textual Relationships through Rewriting}\label{related:rag}

The technique of rephrasing text using Large Language Models (LLMs) to assess the relationship between two documents has gained attention in recent years. 
In the field of Retrieval-Augmented Generation (\cite{lewis2020retrieval}), RaR (Rephrase and Respond) (\cite{deng2023rephrase}) and RAGFusion (\cite{RAGFusion2023}) have shown that using LLMs to rephrase or generate diverse queries from user's query can significantly enhance the performance of information retrieval resulting in better question-answering.

In the context of summarization, there have been studies that measure the factual consistency between the original text and the summarized text by generating questions from the original text and assessing whether the summary can correctly answer these questions (\cite{wang2020asking,kryscinskiFactCC2019}). 
Similarly, studies by \cite{laban-etal-2022-discord} and  \cite{Huang-et-al-NAACL-24} assess summary's diversity of viewpoint from multiple news sources. They create questions from each source and evaluate how well the summary answers these, measuring the summary's coverage of different perspectives. These studies are related to our works in that it rephrases comments into question format and attempts to evaluate the summary beyond quantity measure.

\begin{figure}[t]
\centering
\includegraphics[width=\linewidth]{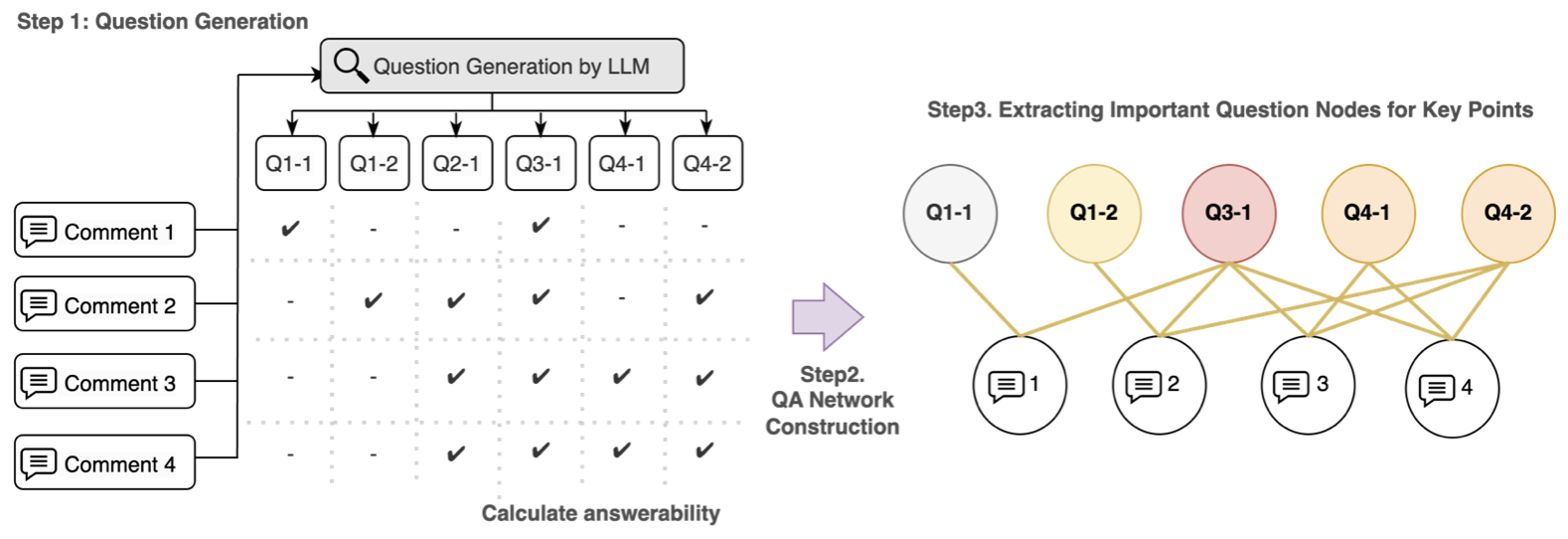}
\caption{Overview of the proposed Question-Answering Network Analysis (QANA) framework for key point analysis. The methodology involves generating questions from comments using Large Language Models (LLMs), constructing a Question-Answering (QA) network based on the answerability of the questions by the arguments, and extracting important question nodes as key points using centrality measures.}
\label{fig:overview}
\end{figure}

\section{Methodology}\label{methodology}

Let $\mathbf{D} = \{d_1, \ldots, d_N\}$ represent the corpus of arguments and $\mathbf{K} = \{k_1, \ldots, k_M\}$ denote the set of key points of the corpus. The objective of this study is to generate the set of key points $\mathbf{K}$ from the corpus of arguments $\mathbf{D}$, which is the same goal as Key Point Generation (KPG) described in Section \ref{kpa}. We propose a novel approach called QANA (Question-Answering Network Analysis) that involves transforming arguments into questions, constructing a QA network, and extracting important question nodes as key points. An overview of the proposed methodology is illustrated in Figure \ref{fig:overview}.

\begin{figure}
\centering
\includegraphics[width=\linewidth]{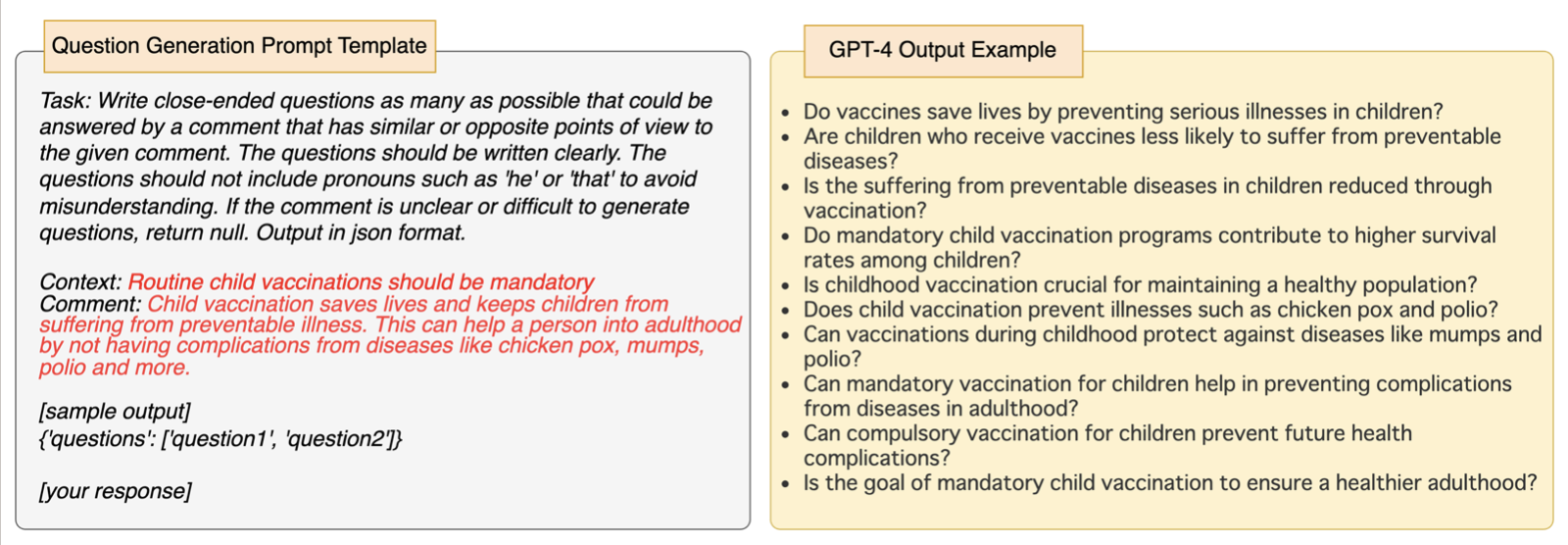}
\caption{Example of argument-to-question transformation using GPT-4 for the topic of mandatory child vaccinations. The prompt template provides guidelines for generating closed-ended questions that capture similar or opposing viewpoints to the given argument. The GPT-4 output successfully extracts multiple perspectives from a single comment, creating a series of questions that represent diverse stances on the topic.}
\label{fig:example}
\end{figure}

\subsection{Argument-to-Question Transformation}

Our method generates a set of questions $\mathbf{Q}i = \{q_{1i}, \ldots, q_{Ki}\}$ from each argument $d_i$. The set of all generated questions is denoted as $\mathbf{Q} = \bigcup_{i=1}^N \mathbf{Q}_i$, and we directly consider these questions as key points, thereby allowing us to formalize $\mathbf{K} \subset \mathbf{Q}$ in our study. Questions are known to serve as an effective framework for fostering discussion and are instrumental in drawing out how other people think (\cite{davis2009tools}).

The specific prompt for question generation and an output example demonstrating the successful separation of multiple perspectives are shown in Figure \ref{fig:example}. By leveraging the power of LLM, transforming arguments into questions enables isolating different viewpoints from a single post, enabling a more granular and comprehensive analysis of the relationships between opinions. 

\subsection{Constructing the QA Network}
After generating questions from the arguments, a QA network is constructed by creating edges between the generated questions and arguments. As demonstrated in the MIRACL benchmark (\cite{10.1162/tacl_a_00595}), the relevance of answer candidates to questions can be efficiently computed using the cosine similarity between their embedding representations. In this study, we leverage this finding and determine the weights of the edges in the QA network based on the cosine similarity between the embeddings of the questions and arguments. The quality of the constructed edges according to embedding models are validated in Section \ref{exp1}.

\begin{equation}
M(d_i, q_j) = \frac{\operatorname{emb}(d_i) \cdot \operatorname{emb}(q_j)}{|\operatorname{emb}(d_i)||\operatorname{emb}(q_j)|}
\end{equation}

where $M(d_i, q_j)$ represents the edge weight between argument $d_i$ and question $q_j$, $\operatorname{emb}(\cdot)$ denotes the embedding function, and $|\cdot|$ represents the L2 norm of the embedding vectors.

\subsection{Extracting Important Question Nodes for Key Points}

By constructing the QA network, we aim to capture the semantic relationships between the generated questions and the original arguments. The resulting network provides a structured representation of the discussions, enabling us to apply network analysis techniques identify important question nodes and extract key points.

For each question $q_i$ in the QA network, a node centrality score (\cite{Lu2016VitalNodes}) $I(q_i)$ is calculated to determine its importance. The choice of centrality measure can be made based on the analyst's interest, allowing for flexibility in the analysis. The top $n$ important questions are denoted as the predicted key points $K_{pred@n}$:

\begin{equation}
K_{pred@n} = \{q_i | I(q_i) \in \text{top-}n\}
\end{equation}

Our top-down approach creates numerous questions and enables the selection of important ones based on the analyst's interest, in contrast to the conventional bottom-up method of clustering arguments. Different centrality measures, such as betweenness (\cite{Freeman1977Set}) or closeness centrality (\cite{FREEMAN1978215}), may be more suitable for ensuring a balanced representation of multiple perspectives in controversial topics.

\section{Experimental Results and Analysis}
In this section, we conduct experiments to address the following research questions:

\begin{itemize}
\item RQ1: How do question type, question generation model, and embedding model affect the quality of the QA network in the Key Point Matching (KPM) setting?
\item RQ2: Do nodes with high centrality scores cover the manually annotated key points?
\end{itemize}

\subsection{Evaluating QA Network Quality in KPM Setting}\label{exp1}

\subsubsection{QA Network Aggregation for KPM}
We evaluate the quality of the QA network by measuring its performance in the KPM task. In the traditional KPM setting, key points are already given, and the task is to match arguments to their corresponding key points. 

Since there are no direct annotations between the generated questions $\mathbf{Q}$ and the manually annotated key points $\mathbf{K}$, a simple aggregation technique is used to assess the connectivity between arguments and key points. For each argument $d_i$, the similarity between the argument and a key point $k_j$ is calculated by averaging the similarity of questions generated from the argument to the key point:

\begin{equation}
M(d_i, k_j) = \frac{1}{|\mathbf{Q}_i|} \sum_{q \in \mathbf{Q}_i} M(q, k_j)
\end{equation}

This evaluation method assumes that if an embedding model can accurately match questions to key points, it is likely to perform well in terms of assessing the answerability of arguments given a question. Furthermore, a high-performing question generation method should create questions that mainly focus on the key points covered by the manual annotations, while a low-performance question method may generate questions focusing on other aspects.

\subsubsection{Dataset and Evaluation Metric}

The ArgKP-2021 dataset (\cite{friedman-etal-2021-overview}) is commonly used for KPM evaluation. It contains arguments on 28 controversial topics, categorized by stance, with each argument annotated as $<$argument, key point, label$>$, where the label indicates whether the argument matches the key point or not.

Performance is evaluated using mean average precision (mAP) of the match between each candidate argument and manually annotated key point pair. Higher matching scores $M(d_i, k_i)$ indicate a stronger match between the argument and the key point. In our experiment, the strict mAP evaluation is used for rigorous assessment, where pairs with missing annotations are considered not matched.

\subsubsection{Comparative Analysis Setup}

The following configurations for QA network generation are compared to understand their impact on KPM performance:

\textbf{Question Type:} Closed-ended, open-ended, and hybrid questions are generated. Closed-ended questions are direct and typically require a "Yes" or "No" answer, providing a specific framework for generating responses. Open-ended questions allow for a wider range of responses, encouraging more detailed and descriptive answers. Hybrid questions involve generating questions without specifically directing them to be either closed or open-ended. In addition to the generated questions, we also use multiple paraphrasing and the original arguments as baselines for comparison.

\textbf{LLM Model:} GPT-3.5 (\cite{openai2023gpt35turbo}; version: "gpt-3.5-turbo-0125") and GPT-4 (\cite{gpt-4-2023-technical}; version: "gpt-4-0125-preview") are used for question generation.

\textbf{Embedding Model:} multilingual-E5 base, large (\cite{wang2024multilingual}) and OpenAI's text-embedding-3-large (\cite{openai_2024_text_embedding_3_large}) are employed.

\textbf{Baseline Methods:} Our approach is compared with the top 10 entries from the Shared Task 2021 (\cite{friedman-etal-2021-overview}), which are trained with ArgKP dataset as described in Section \ref{kpa}, while the proposed framework operates in a zero-shot manner.

\begin{figure}
\centering
\includegraphics[width=\linewidth]{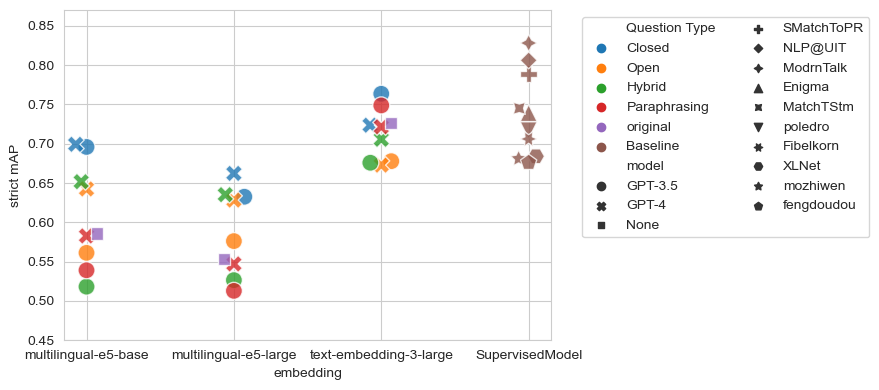}
\caption{Comparison of different question types, language models, and embedding models in terms of their impact on the quality of the Question-Answering (QA) network in the Key Point Matching (KPM) setting. The graph shows the strict mAP scores for various configurations, with the combination of "text-embedding-3-large" embedding and closed questions generated by GPT-4 achieving the highest performance, comparable to state-of-the-art supervised models.}
\label{fig:result1}
\end{figure}

\subsubsection{Results}

Figure~\ref{fig:result1} showcases the evaluation results, demonstrating that the use of text-embedding-3-large achieves performance comparable to traditional supervised learning models, even in a zero-shot scenario. The strong performance of text-embedding-3-large can be attributed to recent advancements in embedding technologies. The proposed approach to KPM reduces computational complexity from quadratic to linear while maintaining comparable performance, as described in Section \ref{kpa}.

The improved performance of closed questions and paraphrasing compared to the original arguments can be attributed to the ensemble effect of increasing the number of variations. However, the impact of these techniques varies depending on the quality of the embedding model used.

For the multi-lingual-e5 embeddings, question generation consistently improves performance, while paraphrasing tends to degrade it. In contrast, when using the text-embedding-3-large model, paraphrasing enhances performance, while question generation methods other than closed questions by GPT-3.5 lead to lower performance compared to the original arguments.

These findings suggest that lower-quality embeddings may not have sufficient language understanding capabilities to accurately match paraphrased comments to key points. In such cases, transforming the text into a question format simplifies the task, leading to better performance. However, as the embedding quality improves, the language understanding ability also increases, enabling the model to correctly match paraphrased texts to key points. Regarding the performance drop observed with non-closed question generation methods when using text-embedding-3-large, it can be inferred that these methods tend to generate a higher proportion of questions that focus on aspects other than the manually created key points, compared to the closed question format.

In conclusion, the text-embedding-3-large model, combined with closed question generation, achieves performance comparable to traditional KPM methods while significantly reducing computational complexity. This approach enables efficient inference that rivals state-of-the-art KPM techniques. Moreover, the results suggest that generating closed questions leads to the creation of a higher proportion of questions that closely align with the manually annotated key points, compared to other question generation methods.

\subsection{Comparing Human-Generated Key Points with High-Centrality Questions}
\subsubsection{Evaluation Protocol}

This subsection addresses the research question, "RQ2: Do nodes with high centrality scores cover the manually annotated key points?"

We investigate the alignment between the important questions identified by centrality measures and the manually created key points in the ArgKP dataset to validate whether the proposed approach effectively captures the key points that human annotators consider crucial.

Previous evaluation methods in related work, such as \cite{li-etal-2023-hear}, utilized metrics like BERTScore or BLEURT to measure the similarity between generated topics and manually annotated key points in a supervised generation setting. However, these methods may not be suitable for our approach due to the inclusion of topic information as context in the generated questions. For example, in the topic "Routine child vaccinations should be mandatory," an example true key point is "The parents and not the state should decide," while our generated question is "Is it preferable for parents, rather than the government, to make health decisions for their children?" Although this question effectively captures the essence of the key point, comparing these questions to the key points using word-level similarity metrics like BERTScore or BLEURT would produce low scores, making the evaluation unfair.

To address this issue, we propose a simpler evaluation method that focuses on the semantic proximity between the correct key points and the generated questions. We determine if the embedding distance between a key point and a generated question, calculated using the text-embedding-3-large model, falls within a predefined threshold that maximizes the difference between the True Positive Rate (TPR) and False Positive Rate (FPR) derived from the binary labels in the ArgKP dataset. This is equivalent to finding the point on the ROC curve closest to the ideal point (0,1), where the false positive rate is 0 and the true positive rate is 1. To avoid including similar questions in the final prediction, duplicate key points within the threshold are considered not in the final prediction.

We calculate the coverage of related key points among the top-n questions with the highest centrality using the following equation:

\begin{equation}
\text{Coverage@n} = \frac{1}{|\mathbf{K}_{true}|} \sum_{k \in \mathbf{K}_{true}} \mathbf{1}\left(\max_{q_i \in \mathbf{K}_{pred@n}} M(k, q_i) \geqq \theta\right)
\end{equation}

where $\text{Coverage@n}$ is the proportion of key points in $\mathbf{K}_{true}$ that have a similarity above the threshold $\theta$ with at least one of the top-n questions $\mathbf{K}_{pred@n}$ based on centrality score $I(q_i)$, $\mathbf{K}_{true}$ is the set of human-generated key points, $|\mathbf{K}_{true}|$ is the total number of key points, and $\mathbf{1}(\cdot)$ is the indicator function.

We evaluate the overall coverage of human-generated key points across different numbers of top questions considered to compare models and question generation methods comprehensively.

\begin{figure}
    \centering
    \includegraphics[width=\linewidth]{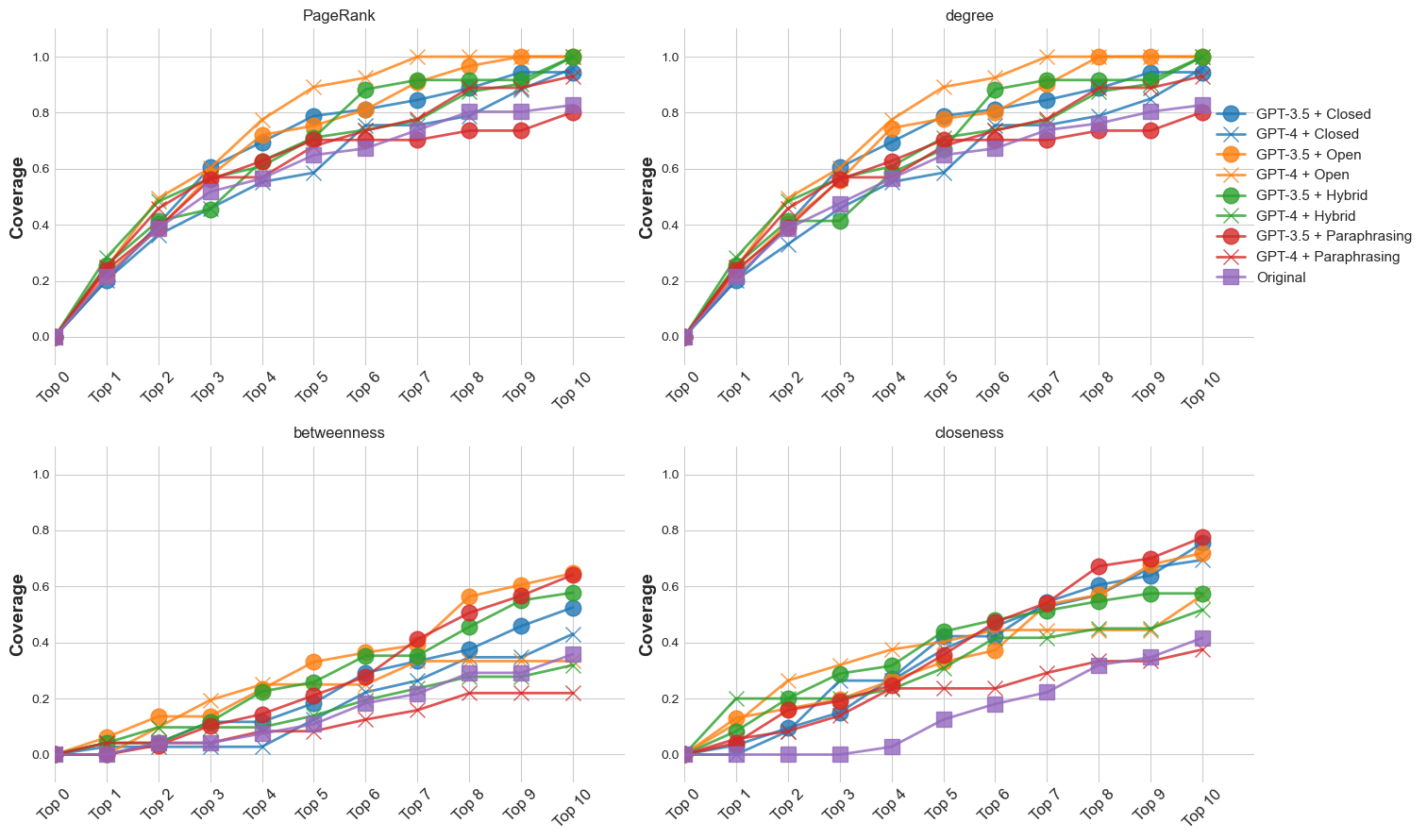}
    \caption{Coverage of human-generated key points by top questions with high centrality scores, comparing different question types and centrality metrics across various model configurations. The graph illustrates the coverage of human-generated key points as the number of top questions considered increases. The combination of GPT-4 for question generation, the "text-embedding-3-large" model for embedding, and PageRank as the centrality metric achieves the highest coverage, with 85\% of key points covered by the top 10 questions.}
    \label{fig:result2}
\end{figure}

\subsubsection{Results}
Figure \ref{fig:result2} presents the coverage of key points among the top 10 questions with the highest centrality scores for various configurations of question types and language models. We used four popular centrality measures: PageRank, degree, betweenness, and closeness.

The results show that there is no significant difference between PageRank and degree centrality or between betweenness and closeness centrality, but the former pair consistently outperforms the latter. These findings suggest that the manually created topics primarily focus on the volume of opinions.

Regarding question generation methods, using questions as a basis for key point generation is more likely to include manually created key points compared to baselines such as the original arguments or paraphrased versions. Interestingly, in contrast to the results from the previous section, open questions performed particularly well in this case. The KPM results in the previous section indicated that closed questions might generate more questions focused on the correct key points. However, the results here show that even when using open questions generated by GPT-4, these key points are still included and are more easily extractable from the generated network using centrality measures.

Overall, the evaluation results demonstrate the effectiveness of the proposed QANA approach in identifying important questions that align with manually annotated key points. However, as we have discussed throughout this paper, it is important to note that achieving performance close to the ground truth topics is not necessarily the most crucial factor from the perspective of fairness and diversity. For reference, we have included examples of generated key points in the Appendix \ref{appendix1}.

\section{Conclusion}
In this study, we proposed a novel opinion mining framework called "Question-Answering Network Analysis" (QANA). The approach involves transforming arguments into questions using Large Language Models (LLMs) and constructing a Question-Answering (QA) network based on each argument's ability to answer those questions. Traditional node centrality measures, such as PageRank or betweenness centrality, are then used to quantify the importance of questions as key points, allowing for interpretations beyond the primarily considered quantitative measures in Key Point Analysis (KPA).

We validated our approach in the KPA setting. In the Key Point Matching (KPM) task, our QA-network achieved performance comparable to traditional supervised learning models, even in a zero-shot setting. We showed that generating closed questions leads to the creation of a higher proportion of questions that closely align with the manually annotated key points. Our approach also reduced the computational cost of matching key points and arguments from quadratic to linear. In the Key Point Generation (KPG) task, questions with high PageRank or degree centrality scores covered almost all of the manually crafted key points.

The primary contribution of this study is the proposal of a flexible framework that allows analysts to assess the importance of key points from various aspects according to their interests. This approach provides a valuable tool for opinion mining in the broadcasting field, where impartiality and fairness are crucial considerations.

There are several directions for future research to further enhance the proposed framework. While this study focused on the importance of questions based on centrality, it would also be interesting to take into account the diversity of answers, such as the ratio of Yes or No responses in closed questions or the diversity of answers in open questions. Such questions, where opinions are divided, are particularly important in terms of fairness. Another potential avenue for future research is to investigate the effectiveness of QANA on real-world data. Our question rephrasing technique may perform well when comments are mostly impressions and not explicitly opinions, and background information is necessary for understanding. Validating the performance of QANA on such datasets could further demonstrate its utility in from traditional approachs in real-world scenarios. Additionally, examining the relationship between each centrality measures could provide further insights. Further research on debate intensity, validation on real-world data and additional centrality measures will enhance the framework's applicability and impact in opinion mining.

\bibliography{colm2024_conference}
\bibliographystyle{colm2024_conference}

\appendix
\section{Appendix}\label{appendix1}
Table \ref{table:child_vaccinations_analysis} shows the example of questions generated using GPT-4 with Top-5 highest centrality, based on opposition viewpoints regarding the topic "Social media platforms should be regulated by the government." 

\begin{table}[ht]
\centering
\begin{tabularx}{\textwidth}{|l|X|}
\hline
\textbf{Annotated Keypoint} &
$\cdot$ Social media regulation is not effective  \newline
$\cdot$ Social media regulation harms privacy \newline
$\cdot$ Social media regulation harm freedom of speech and other democratic rights \newline
$\cdot$ The government should not intervene in the affairs of a private company \newline
$\cdot$ Social media regulation can lead to political abuses by the government 
\\ \hline
\textbf{PageRank} & 
$\cdot$ Why should social media platforms remain free from government regulation? \newline
$\cdot$ How could freedom of expression be affected if governments start regulating social media content? \newline
$\cdot$ Can the concept of privacy on social media coexist with government regulation? \newline
$\cdot$ Could there be a conflict of interest if the government, which is often a subject of discussion on these platforms, is in charge of regulating them? \newline
$\cdot$ Why is it important to keep some posts hidden from the government on social media platforms? 
\\ \hline
\textbf{Degree} & 
$\cdot$ Why should social media platforms remain free from government regulation?\newline
$\cdot$ How could freedom of expression be affected if governments start regulating social media content?\newline
$\cdot$ Can the concept of privacy on social media coexist with government regulation?\newline
$\cdot$ Could there be a conflict of interest if the government, which is often a subject of discussion on these platforms, is in charge of regulating them?\newline
$\cdot$ Why is it important to keep some posts hidden from the government on social media platforms?
\\ \hline
\textbf{Betweenness} & 
$\cdot$ In what ways does social media foster community and enable connections across distances?\newline
$\cdot$ What are the key differences between platforms used for fun and those used for professional or informational purposes?\newline
$\cdot$ What safeguards should be implemented to ensure that the reporting system does not get misused by individuals with malicious intent?\newline
$\cdot$ In what ways does social media provide a platform for emerging artists and creators?\newline
$\cdot$ How does social media contribute to positive mental health outcomes?
\\ \hline
\textbf{Closeness} & 
$\cdot$ In what ways does social media foster community and enable connections across distances?\newline
$\cdot$ In what ways does social media provide a platform for emerging artists and creators?\newline
$\cdot$ What are the key differences between platforms used for fun and those used for professional or informational purposes?
$\cdot$ How does social media contribute to positive mental health outcomes?\newline
$\cdot$ Can social media be considered a vital tool for learning and gaining new perspectives?
\\ \hline
\end{tabularx}
\caption{This table presents a collection of open questions generated using GPT-4, based on opposition viewpoints regarding the topic "Social media platforms should be regulated by the government."}
\label{table:child_vaccinations_analysis}
\end{table}

\end{document}